\documentclass[sigconf]{acmart}

\AtBeginDocument{%
  \providecommand\BibTeX{{%
    \normalfont B\kern-0.5em{\scshape i\kern-0.25em b}\kern-0.8em\TeX}}}

\usepackage{multirow}
\usepackage{url}
\usepackage{siunitx}

\copyrightyear{2021}
\acmYear{2021}
\setcopyright{rightsretained}
\acmConference[GECCO '21 Companion]{2021 Genetic and Evolutionary Computation Conference Companion}{July 10--14, 2021}{Lille, France}
\acmBooktitle{2021 Genetic and Evolutionary Computation Conference Companion (GECCO '21 Companion), July 10--14, 2021, Lille, France}\acmDOI{10.1145/3449726.3459490}
\acmISBN{978-1-4503-8351-6/21/07}

\begin{document}

\title{ARCH-Elites: Quality-Diversity for Urban Design}
\author{Theodoros Galanos}
\affiliation{%
  \institution{University of Malta}
  \city{Sliema}
  \country{Malta}}
\email{theodoros.galanos@um.edu.mt}

\author{Antonios Liapis}
\affiliation{%
  \institution{University of Malta}
  \city{Sliema}
  \country{Malta}}
\email{antonios.liapis@um.edu.mt}

\author{Georgios N. Yannakakis}
\affiliation{%
  \institution{University of Malta}
  \city{Sliema}
  \country{Malta}}
\email{georgios.yannakakis@um.edu.mt}

\author{Reinhard Koenig}
\affiliation{%
  \institution{Bauhaus-University Weimar, Austrian Institute of Technology}
  \city{Weimar}
  \country{Germany}}
\email{reinhard.koenig@uni-weimar.de}

\begin{abstract} 
This paper introduces ARCH-Elites, a MAP-Elites implementation that can reconfigure large-scale urban layouts at real-world locations via a pre-trained surrogate model instead of costly simulations. In a series of experiments, we generate novel urban designs for two real-world locations in Boston, Massachusetts. Combining the exploration of a possibility space with real-time performance evaluation creates a powerful new paradigm for architectural generative design that can extract and articulate design intelligence.
\end{abstract}



\maketitle

\section{ARCH-Elites}\label{sec:introduction}

\begin{figure}
\centering
\includegraphics[width=\linewidth]{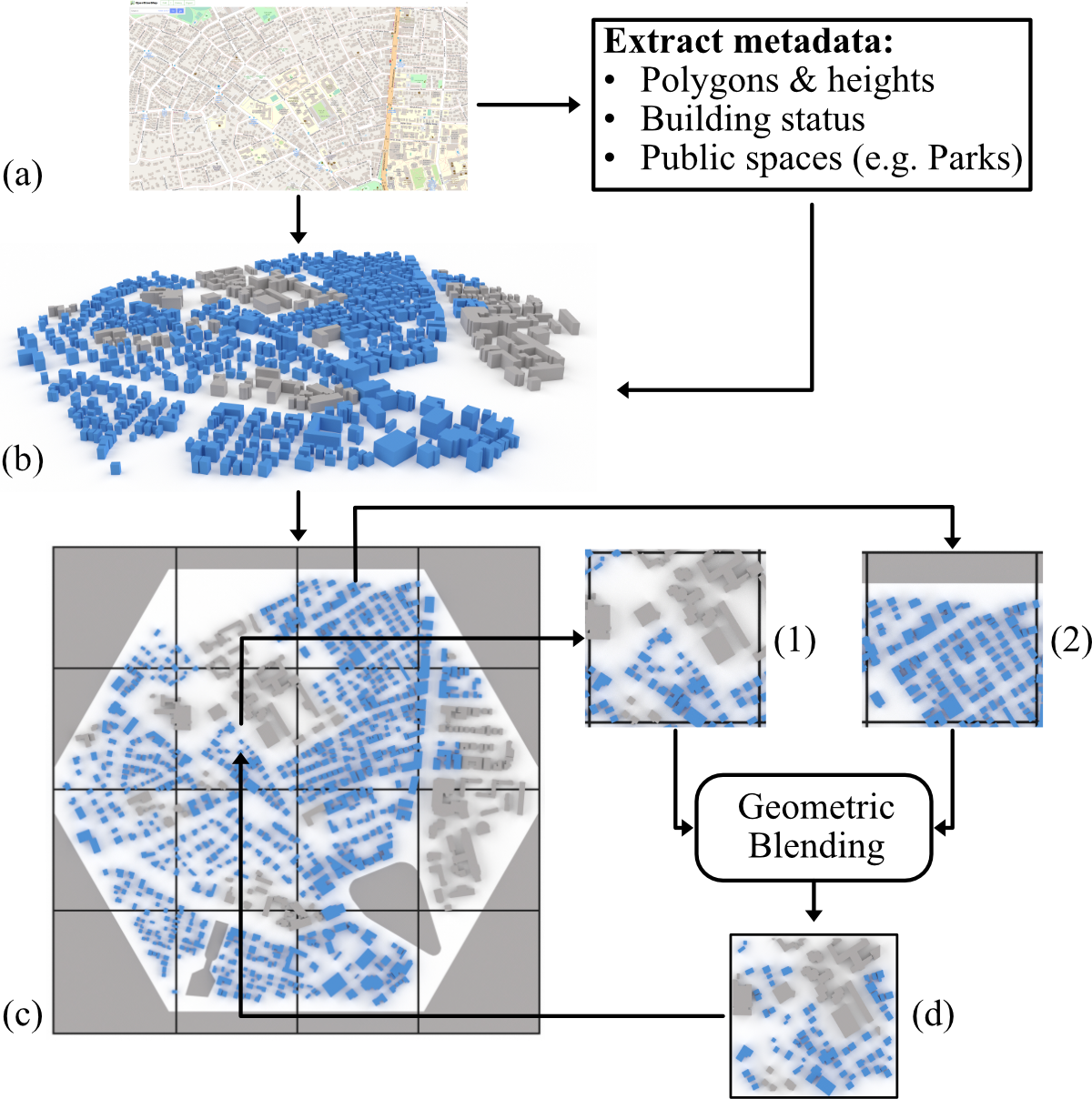}
\caption{Evolution in ARCH-Elites: A geographic location is selected by the user (a) and building geometries and metadata are extracted (b). The location is then split into cells (c); two cells are selected and geometric blending takes place (d).}
\label{fig:evolution}
\end{figure}

ARCH-Elites is a MAP-Elites \cite{mouret2015illuminating} implementation that can reconfigure large-scale urban layouts at real-world locations via a pre-trained surrogate model. We argue that Quality-Diversity (QD) search \cite{gravina2019pcgqd,pugh2016quality} is an ideal design exploration and generation process for architectural and urban design problems, and apply it in this paper to answer the question: \emph{``What is the most dense urban design within a specific range of comfortable and dangerous conditions?''}.

In this implementation of ARCH-Elites, the genome of each layout is a collection of building polygons (a set of 2D coordinates) and their heights. A binary value ($f$) is added to each polygon's genome to indicate whether it can be modified or not (i.e. ``static'').

The \emph{behavioral space} is the city's resilience to extreme weather events, evaluated through the pre-trained surrogate model \textit{InFraRed}. Wind flow around buildings is evaluated under storm wind conditions (i.e. 17 m/s). The resulting dimensions for the MAP-Elites feature map are (a) the percentage of open space within \textit{comfortable sitting conditions} ($B_c$) and (b) the total open space area that exhibits \textit{dangerous conditions} ($B_d$). Winds less than 6 m/s are considered comfortable for sitting and winds higher than 15 m/s considered dangerous for pedestrians \cite{lawson1978wind}. The metrics $B_c$ and $B_d$ are evaluated based on 4 wind directions multiplied by the total number of hours in the year during which a wind in that direction blows on the tested location, based on real-world data.

The \emph{quality} dimension in ARCH-Elites is the Floor Space Index (FSI), defined as the ratio of the gross floor area divided by the total plot area. The number of floors is calculated based on a constant floor height (4m). Finally, each building's gross floor area is calculated by multiplying the number of floors with its footprint. 

The evolutionary process begins by selecting an elite location from the feature map, following a uniform random distribution. In the first iteration, only the initial real-world location is in the feature map, so it is always selected. 
The selected individual is then transformed through a number of custom operators. The first operator selects a pair of cells at random and buildings are exchanged between them. A percentage of buildings are removed from the first cell (between $10\%$ and $50\%$) and all buildings of the second cell are overlayed with the remaining buildings of the first cell. The buildings of the second cell that are not intersecting those of the first cell are copied to the first cell. This is depicted in Fig.~\ref{fig:evolution} as ``geometric blending''. If no changes are made from this process, e.g. if all overlayed buildings were intersecting, another pair of cells is selected until the blending is successful. The resulting cell then undergoes a polynomial bounded mutation of its building heights, inspired by \cite{deb2002nsga}. After geometric blending and height mutations, the changed (first) cell is re-added to the location at its original coordinates (see Fig.~\ref{fig:evolution}). 
For each selected individual, this process is repeated 5 times and thus produces a new individual with up to 5 cells that are different from the original. The mutated individual is then copied, and its copy is mutated another 5 times to create two offspring with different degrees of change. The new individuals' behavioral dimensions are calculated via InFraRed and each new individual replaces a less fit individual in the feature map.

\section{Case Study: Boston}\label{sec:experimental_setup}

We conducted two real-world experiments using ARCH-Elites, for two different locations within the city of Boston, Massachusetts: an area around the Harvard University's campus and an area around the MIT campus. University buildings, public buildings and public spaces were encoded as static within the genotype (gray in Fig.~\ref{fig:evolution}) and did not change during evolution. 
The resolution of the feature map is 20 $\times$ 20 bins, with $B_c\in[0.7, 0.9]$ for Harvard, $B_c\in[0.5, 0.7]$ for MIT, and $B_d\in[0, 8000]$ in both locations. The height of buildings is capped between 4 and 100 meters.
Each QD run performs $2,000$ selections (i.e. $4,000$ evaluations). We report the maximum fitness in the run's archive, coverage \cite{mouret2015illuminating} and QD-score \cite{pugh2016quality}. Ten QD runs are performed per location, and all elites of the 10 evolutionary runs are accumulated to assess the patterns of final evolved cities.

\begin{table}
\small
\centering
\caption{Results at the end of 10 evolutionary runs.}
\label{tab:results}
\begin{tabular}{|l||c|c|c|c|}
\hline 
{Performance Metrics} & \multicolumn{2}{c|}{Harvard} & \multicolumn{2}{c|}{MIT} \\ 
\hline
Coverage & \multicolumn{2}{c|}{0.46} & \multicolumn{2}{c|}{0.29} \\
Maximum Fitness & \multicolumn{2}{c|}{1.01} & \multicolumn{2}{c|}{1.63} \\
QD-score & \multicolumn{2}{c|}{175} & \multicolumn{2}{c|}{175} \\
\hline \hline
\multicolumn{5}{|c|}{Mean values of all final elites} \\ \hline
& \multicolumn{2}{c|}{Harvard} & \multicolumn{2}{c|}{MIT} \\ 
\cline{2-5} 
{Urban Characteristics} & {Evolved} & {Original} & {Evolved} & {Original}\\ 
\hline 
Building height ($m$) & 22.4 & 17.1 & 20.6 & 17.9 \\
Num. buildings & 678 & 781 & 721 & 980\\
Open Space Ratio & 0.82 & 0.67 & 0.56 & 0.46 \\
Floor Space Index & 0.91 & 0.81 & 1.4 & 1.49  \\
Comfortable space ($\%$) & 77  & 84 & 54 & 60 \\
Dangerous space ($m^2$) & 2303 & 218 & 2799 & 357 \\
\hline
\end{tabular}
\end{table}

The evolutionary process does not seem to improve the maximum fitness substantially; for the MIT area the maximum fitness increases only by $20\%$ for the Harvard area and by $2.5\%$ for the MIT area, compared to the initial population. The fittest elite had an FSI over $8\%$ above the original MIT layout, and over $19\%$ above the original Harvard layout (see Table \ref{tab:results}). This indicates a difficulty of the generative process to find designs that are much denser than the initial MIT location, while this seems possible for Harvard. The dense layout of the MIT area (with an open space ratio of $46\%$, compared to $67\%$ for Harvard) means that no additional buildings are likely to be added. A large portion of the footprint is static for the MIT area ($39\%$ versus $10\%$ for Harvard), exacerbating the problem.

Based on Table~\ref{tab:results}, most elites have a higher FSI than the real-world location for Harvard and a lower FSI than the real-world location for MIT. Only $5\%$ of elites in the MIT area had more buildings than the original layout, versus $22\%$ in the Harvard area. In terms of the main behavioral characteristics, only $5\%$ of evolved cities have more comfortable space and $97\%$ have a larger dangerous areas for Harvard, and $0.3\%$ have more comfortable space and $96\%$ more dangerous space for MIT. The fact that most elites have more open spaces leads to an increase in dangerous areas, while when elites have more buildings more wind tunnels are likely to occur and the comfortable area drops. This observation illustrates the kind of complex challenge that both human urban designers and evolutionary algorithms have to face.

\section{Conclusion}\label{sec:conclusion}

This paper views urban design as a well-fitted domain for QD search and tests how MAP-Elites can be applied for exploring the trade-offs between denser cities and dangerous conditions for pedestrians in storm weather conditions. Taking advantage of a pre-trained surrogate model, we could perform evolutionary runs of $4,000$ evaluations within a few hours, compared to the week-long simulation time required for a single wind comfort study. Moreover, the feature maps produced by the QD runs can be particularly illustrative for an urban designer and inspire further human design iterations.

\begin{acks}
This project has received funding from the EU’s Horizon 2020 programme under grant agreement No 952002.
\end{acks}

\bibliographystyle{ACM-Reference-Format}
\bibliography{poster}


\begin{thebibliography}{5}


\ifx \showCODEN    \undefined \def \showCODEN     #1{\unskip}     \fi
\ifx \showDOI      \undefined \def \showDOI       #1{#1}\fi
\ifx \showISBNx    \undefined \def \showISBNx     #1{\unskip}     \fi
\ifx \showISBNxiii \undefined \def \showISBNxiii  #1{\unskip}     \fi
\ifx \showISSN     \undefined \def \showISSN      #1{\unskip}     \fi
\ifx \showLCCN     \undefined \def \showLCCN      #1{\unskip}     \fi
\ifx \shownote     \undefined \def \shownote      #1{#1}          \fi
\ifx \showarticletitle \undefined \def \showarticletitle #1{#1}   \fi
\ifx \showURL      \undefined \def \showURL       {\relax}        \fi
\providecommand\bibfield[2]{#2}
\providecommand\bibinfo[2]{#2}
\providecommand\natexlab[1]{#1}
\providecommand\showeprint[2][]{arXiv:#2}

\bibitem[\protect\citeauthoryear{Deb, Pratap, Agarwal, and {Meyarivan}}{Deb
  et~al\mbox{.}}{2002}]%
        {deb2002nsga}
\bibfield{author}{\bibinfo{person}{Kalyanmoy Deb}, \bibinfo{person}{Amrit
  Pratap}, \bibinfo{person}{Sameer Agarwal}, {and} \bibinfo{person}{T.
  {Meyarivan}}.} \bibinfo{year}{2002}\natexlab{}.
\newblock \showarticletitle{A fast and elitist multiobjective genetic
  algorithm: NSGA-II}.
\newblock \bibinfo{journal}{\emph{IEEE Transactions on Evolutionary
  Computation}} \bibinfo{volume}{6}, \bibinfo{number}{2}
  (\bibinfo{year}{2002}), \bibinfo{pages}{182--197}.
\newblock


\bibitem[\protect\citeauthoryear{Gravina, Khalifa, Liapis, Togelius, and
  Yannakakis}{Gravina et~al\mbox{.}}{2019}]%
        {gravina2019pcgqd}
\bibfield{author}{\bibinfo{person}{Daniele Gravina}, \bibinfo{person}{Ahmed
  Khalifa}, \bibinfo{person}{Antonios Liapis}, \bibinfo{person}{Julian
  Togelius}, {and} \bibinfo{person}{Georgios~N. Yannakakis}.}
  \bibinfo{year}{2019}\natexlab{}.
\newblock \showarticletitle{Procedural Content Generation through
  Quality-Diversity}. In \bibinfo{booktitle}{\emph{Proceedings of the IEEE
  Conference on Games}}.
\newblock


\bibitem[\protect\citeauthoryear{Lawson}{Lawson}{1978}]%
        {lawson1978wind}
\bibfield{author}{\bibinfo{person}{T.~V. Lawson}.}
  \bibinfo{year}{1978}\natexlab{}.
\newblock \showarticletitle{The wind content of the built environment}.
\newblock \bibinfo{journal}{\emph{Journal of Wind Engineering and Industrial
  Aerodynamics}} \bibinfo{volume}{3}, \bibinfo{number}{2}
  (\bibinfo{year}{1978}), \bibinfo{pages}{93--105}.
\newblock


\bibitem[\protect\citeauthoryear{Mouret and Clune}{Mouret and Clune}{2015}]%
        {mouret2015illuminating}
\bibfield{author}{\bibinfo{person}{Jean-Baptiste Mouret} {and}
  \bibinfo{person}{Jeff Clune}.} \bibinfo{year}{2015}\natexlab{}.
\newblock \bibinfo{title}{Illuminating search spaces by mapping elites}.
\newblock
\newblock
\showeprint[arxiv]{1504.04909}~[cs.AI]


\bibitem[\protect\citeauthoryear{Pugh, Soros, and Stanley}{Pugh
  et~al\mbox{.}}{2016}]%
        {pugh2016quality}
\bibfield{author}{\bibinfo{person}{Justin~K Pugh}, \bibinfo{person}{Lisa~B
  Soros}, {and} \bibinfo{person}{Kenneth~O Stanley}.}
  \bibinfo{year}{2016}\natexlab{}.
\newblock \showarticletitle{Quality Diversity: A New Frontier for Evolutionary
  Computation}.
\newblock \bibinfo{journal}{\emph{Frontiers in Robotics and AI}}
  \bibinfo{volume}{3} (\bibinfo{year}{2016}).
\newblock


\end{thebibliography}

\end{document}